\documentclass[review]{elsarticle}

\usepackage{lineno,hyperref}

\usepackage{graphics}
\graphicspath{}

\biboptions{sort&compress}
\usepackage{amsmath}
\usepackage{verbatim}
\usepackage{cleveref}

\usepackage{algorithm}  
\usepackage{algpseudocode}  
\usepackage{amsmath}  
\usepackage{caption} 
\usepackage{array}
\usepackage{txfonts} 
\usepackage{ragged2e}
\usepackage{url}

\usepackage{array} 

\usepackage{booktabs} 

\usepackage{longtable}

\usepackage{bm}

\usepackage{setspace} 

\usepackage{natbib} 

\begin{document}

\captionsetup[figure]{labelfont={bf},name={Fig.},labelsep=period}
\captionsetup[table]{labelfont={bf},name={Table.},labelsep=period}
\begin{frontmatter}

\title{Cross-Modal Image Fusion Theory Guided by Subjective Visual Attention}
\author{Aiqing Fang$^a$, Xinbo Zhao$^a$, Yanning Zhang$^a$}%
\address{
	\justifying\let\raggedright\justifying
	$^a$National Engineering Laboratory for Integrated Aero-Space-Ground-Ocean Big Data Application Technology, School of Computer Science and Engineering, Northwestern Polytechnical University, Xi’an 710072, China

}



\cortext[mycorrespondingauthor]{Corresponding author}
\ead{xbozhao@nwpu.edu.cn (X. Zhao), aiqingf@mail.nwpu.edu.cn (A. Fang), ynzhang@nwpu.edu.cn}

\begin{abstract}
The human visual perception system has very strong robustness and contextual awareness in a variety of image processing tasks. This robustness and the perception ability of contextual awareness is closely related to the characteristics of multi-task auxiliary learning and subjective attention of the human visual perception system. In order to improve the robustness and contextual awareness of image fusion tasks, we proposed a multi-task auxiliary learning image fusion theory guided by subjective attention. The image fusion theory effectively unifies the subjective task intention and prior knowledge of human brain. In order to achieve our proposed image fusion theory, we first analyze the mechanism of multi-task auxiliary learning, build a multi-task auxiliary learning network. Secondly, based on the human visual attention perception mechanism, we introduce the human visual attention network guided by subjective tasks on the basis of the multi-task auxiliary learning network. The subjective intention is introduced by the subjective attention task model, so that the network can fuse images according to the subjective intention. Finally, in order to verify the superiority of our image fusion theory, we carried out experiments on the combined vision system image data set, and the infrared and visible image data set for experimental verification. The experimental results demonstrate the superiority of our fusion theory over state-of-arts in contextual awareness and robustness.
\end{abstract}

\end{frontmatter}

\section{Introduction}
In the task of image fusion, the robustness and the perception ability of dynamic contextual awareness have been the bottleneck of the application and promotion of existing image fusion technology, while the task of multi-source image fusion has strong robustness and the perception ability of contextual awareness. The research of cognitive psychology and neurobiology \cite{ParisiGermanI2019Cllw,GuangYang2009Smds} shows that human beings have the ability of fast and continuous learning, which is closely related to the multi-task assisted learning mechanism guided by working memory and the characteristics of subjective attention. It is based on these two characteristics of human brain that makes human beings have strong robustness and dynamic contextual perception ability in various computer vision fields. In recent years, many image fusion algorithms \cite{Bavirisetti2016Two,Lahoud2019FastZERO,Liu2017InfraredCNN,MaFusionGAN} have been proposed inspired by biological characteristics, but visual attention is rarely studied by existing image fusion algorithms. The study of visual attention is more based on the difference of contrast, brightness and other information of the data itself to obtain the significant characteristic map \cite{Ma2017InfraredWLS,Bavirisetti2016Two,Lahoud2019FastZERO,Liu2017InfraredJSR-SD}, without considering the relationship between the top-down \cite{Ma2017Multi} task guided subjective visual attention characteristics and cross-modal image fusion task. In the task of image fusion, the existing image fusion algorithms, especially deep learning methods that  lacks of ground truth labels, will carry out image fusion regardless of whether the image information is helpful to humans subjective intention or not. The main reasons include the following three aspects. First of all, the deep learning method has a serious dependence on the objective loss function \cite{fang2019crossmodal}, but at present, the complete image quality objective loss function has not been found for the image that lacks ground truth labels \cite{fang2019crossmodal}. Secondly, the existing image fusion theory is more from the task of image fusion to improve the image quality. But image fusion is not only for humans subjective aesthetic needs, but also for the purpose of assisting human beings to complete specific tasks quickly and accurately through the complementary advantages of different image data in practical application tasks. Finally, in some image fusion tasks that need to ensure that human visual attention is not distracted (image fusion task of visual system in the aviation field, or infrared and visible image fusion task in military war, etc.), higher requirements are put forward for the improvement of image fusion theory. For example, in the aviation combined vision system (CVS) image fusion task, by the fusion of the enhanced vision system (EVS) image and the synthetic vision system (SVS) image, pilots can effectively improve the contextual perception ability of airport runway information. However, the CVS image fusion technology needs to meet the following two conditions: on the one hand, due to the limitations of airborne hardware, the real-time performance of image fusion has high requirements; on the other hand, the effect of image fusion must consider the characteristics of human attention to reduce the impact of scattered fusion information on human attention. In this case, the existing image fusion theory is no longer applicable. At the same time, when dealing with new tasks, human beings tend to work according to the guidance of subjective intention. 

To solve these problems, we proposed an image fusion theory based on human cognitive psychology. This theory effectively combines humans subjective attention with image fusion task, reduces the amount of image fusion data, and improves the contextual awareness perception ability of image fusion. Our image fusion theory is not involved in the research process of the existing image fusion algorithm. Because our image fusion theory is based on the subjective attention guidance when people deal with specific tasks, our image fusion effect and the traditional image fusion effect will be different in the form of expression, But our fusion effect is in line with human visual characteristics, and it is helpful for the tasks that human beings are currently dealing with.

\begin{figure}[ht]
	\centering

	\includegraphics[scale=3.5,width=1\textwidth]{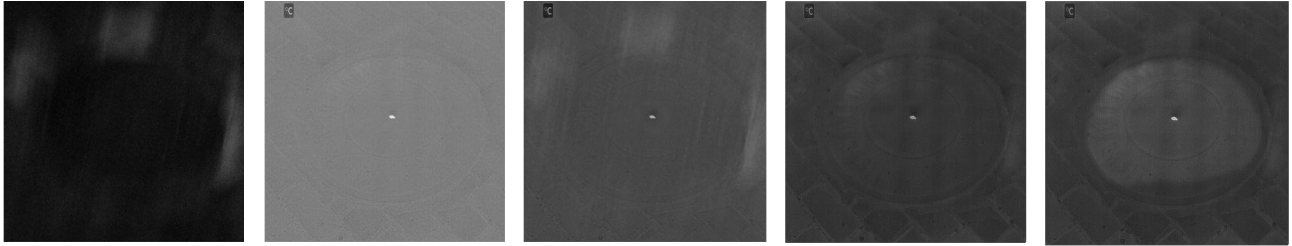}

			\caption{ Schematic illustration of infrared and visible image fusion. From left to right: the visible image, the infrared image, the third is the fusion result of traditional image fusion mathod, the last two image are the fusion results of our image fusion theory. Our method has a good fusion effect for image details, and the fusion effect is more coincident with human visual perception mechanism. At the same time, our image fusion theory highlights the region of human concern. 
		}
		\label{summary}
	\end{figure}

In order to demonstrate the superiority of our image fusion theory over the existing mainstream algorithmic framework, we give a representative example in \ref{summary} . In order to highlight the advantages of our algorithm framework, as shown in \ref{summary}, we use the infrared and visible image data set to do qualitative comparison experiments. In order to facilitate the subjective visual contrast of image quality, we have also carried out image fusion operation for non visual areas of concern. In practical application, we have only enhanced fusion for areas of human subjective visual concern.

The main contributions of our work include the following three points:

First of all, we analyze the subjective attention characteristics of the human visual system, proposed an image fusion theory guided by human subjective visual attention mechanism, and on this basis, a cross modal subjective visual attention detection method is proposed.

Secondly, we analyze the characteristics of human brain using prior knowledge to auxiliary learning in new tasks, and the theory of cross-modal image fusion optimization based on multi-task auxiliary learning is proposed.

Then, aiming at the problem of global feature missing in the process of image fusion, we proposed an image fusion theory combining local feature and global feature.

Finally, based on the theory of multi-task auxiliary learning image fusion guided by subjective attention, we proposed an unsupervised learning network image fusion framework. The results show that our image fusion theory has stronger robustness and contextual awareness than the existing image fusion theory.

\section{Related work}
In this chapter, we will review the visual attention mechanism related to image fusion tasks, the mechanism of multi-task auxiliary learning, the related research work related to feature presentation, and the optimization theory of multi-task auxiliary image fusion tasks guided by human visual attention is proposed.

\subsection{Visual attention}
\label{method}
As a manifestation of human intention, visual attention is widely used in target detection \cite{WangW.2018SODD}, object segmentation \cite{WangWenguan2015Sgvo,WenguanWang2018SVOS}. According to different sources, the existing visual attention can be divided into two types: bottom-up \cite{TheeuwesJ.2005Tabc,YanYin2018Bsat} attention model and top-down \cite{TheeuwesJ.2005Tabc,Ma2017Multi,YanYin2018Bsat} visual attention model. The bottom-up unconscious visual attention is also called significant attention, which is mainly based on the difference of underlying features such as brightness, contrast and edge of image data itself. This attention does not need active intervention and has nothing to do with humans subjective intention. The top-down conscious attention is also called focused attention, which is related to humans subjective intention or task. In the task of image fusion, the current research on visual attention is mainly the method of visual saliency map and weighted least square optimization proposed by \cite{Zhang2015A} for the task of infrared and visible image fusion, which mainly combines non-subsampled Shearlet transformation and visual saliency map for image fusion. Later, although some image fusion algorithms \cite{Lahoud2019FastZERO,ZhangInfrared,Bavirisetti2016Two,ZhangInfrared,Liu2017InfraredJSR-SD,Ma2017InfraredWLS} are proposed, the visual attention involved in these image fusion algorithms is only the use of the saliency feature map of the underlying data, which is no different from the method proposed by \cite{Zhang2015A} in principle. At the same time, the research results of cognitive psychology also show that \cite{Treisman1980A}, human visual attention has the characteristics of selection, which includes the selection of feature weight and feature region. In the aspect of feature weight selection, we filter according to the importance of features. For the effective feature channel \cite{HuJie2019SN}, it gives more weight, and for the invalid feature channel, it gives less weight. The region selection of visual attention \cite{JaderbergM.2015Stn,WangX.2018NNN} is mainly determined by the weighting of all pixel position features rather than the pixel itself. Although these two characteristics have been widely used in many fields of computer vision, there are few related research results in the field of image fusion. In the field of image fusion, channel attention module is used for multi-focus image fusion task for the first time \cite{YanXiang2018UDMI,fang2019crossmodal}. In this paper, we use a top-down task-guided attention model to introduce feature selection and region selection characteristics of visual attention in the process of cross-modal subjective attention detection.

\subsection{Multi-task auxiliary learning}
With the development of deep learning technology, some unsupervised deep learning methods are proposed for the task of cross-modal image fusion. For example, To solve the problem of insufficient image feature extraction, Li \cite{Li2018DenseFuse} proposed a densefuse deep learning network framework for infrared and visible image fusion task. This method uses structural similarity index (SSIM) \cite{1284395} and pixel loss as loss functions to reconstruct single-modal image, and the fusion criteria are only used in the test phase. Although the deep learning method is used in this method, there are two problems. Firstly, the fusion criterion is not added to the model training, which makes the network unable to learn the fusion weight of cross-modal images; secondly, only SSIM and pixel objective function are used as loss function in single-modal image reconstruction. Because of the visual masking effect of human visual perception system, SSIM image evaluation index is not enough to represent the image quality when the image quality degradation is serious. Aiming at the problems of traditional image fusion methods and deep learning methods, Li \cite{Li_2018DL,Lahoud2019FastZERO} proposed a deep learning image fusion method based on knowledge complementation between traditional method and deep learning method, the deep learning features are extracted through VGG19 model, and weighted average and maximum methods are still used in the fusion criteria. In order to improve the generality of image fusion algorithm, Zhang \cite{ZhangYu2020IAgi} proposed a general image fusion framework (IFCNN) based on the full convolution neural network. The method performs supervised learning training on multi-focus data sets, then changes image fusion criteria according to different image fusion tasks, and applies the pretraining weight directly to cross-modal image fusion tasks. Finally, the result of fusion is enhanced by the CLAHE algorithm. Although the method has achieved some results, but for different image fusion tasks, image noise data distribution is different, especially in the cross-modal image data set. Therefore, the weights trained on a single dataset cannot effectively simulate multiple different data noise distributions. As the same time, these image fusion algorithms focus on the feature extraction and combination through the network design, and they do not fundamentally study the learning optimization problem caused by the imperfect objective loss function of image fusion task. In view of this problem, although there are some image quality evaluation methods based on generative network \cite{LinK.-Y.2018HNIQ} and deep learning \cite{KimJongyoo2017Dbiq} in the field of image quality evaluation, they are still quite different from human subjective evaluation. According to the relevant research of cognitive psychology \cite{MillerCPortex,LiuDing2014Mpad}, the human brain can reason and explore new target tasks according to working memory, so as to solve the common and characteristic characteristics of new tasks, which is the multi-task assistant learning characteristics of the human visual perception system. Working memory here is the transcendental knowledge learned by multitasking. Although the mechanism of multi-task auxiliary learning has achieved remarkable results in the computer vision tasks such asimage segmentation \cite{KendallAlex2018MLUU}, head pose detection \cite{WangHaofan2019Hccf}, no relevant research results have been found in the field of image fusion. This mechanism provides a new idea for image fusion theory. In the task of image fusion, by introducing the mechanism of multi-task auxiliary learning, the fusion process can be more in line with the mechanism of human visual perception; on the other hand, by introducing the mechanism of multi-task assisted learning, we can avoid the impact of imperfect target loss on the quality of image fusion, which provides a new idea for image fusion. Inspired by this, based on the previous research on the mechanism of multi-task auxiliary learning \cite{fang2019crossmodal}, we introduce image enhancement task and subjective attention detection task into the task of cross-modal image fusion. 

\subsection{Global and local features}
In the existing image fusion theory, in order to improve the quality of image fusion, people will combine the traditional image processing method with the deep learning method \cite{Lahoud2019FastZERO,Li_2018DL,Liu2017InfraredCNN,Yu2017medicalCNN}, because on the one hand, this method can overcome the problem that the traditional algorithm is not robust to extract features, on the other hand, it can also alleviate the problem of global feature missing in the end-to-end image fusion theory to a certain extent \cite{Lahoud2019FastZERO,Li_2018DL,Liu2017InfraredCNN,Yu2017medicalCNN,PrabhakarK.Ram2017DADU,MaBoyuan2019SAUD}. Although in the end-to-end image fusion network, \cite{ZhangYu2020IAgi,Li2018DenseFuse} proposes to optimize the convolution weight of the last layer of the pre-trained network as the initial weight of the first layer of convolution, so as to introduce global features, the extraction of high-level semantic features is not determined by a certain layer of convolution weight, but by the combination of multi-level convolution weight and pooling. This method only takes the high-level weight of the pre-trained model as the weight of the first layer of the image fusion network. Due to the lack of the pooling layer, the extracted features are still local features. As the same time, although many methods about global and local feature extraction (Pyramid \cite{Burt1987TheLP}, Multi-Scale Transform \cite{Liu2015ALPSR}, Wavelet Transform \cite{Chipman1995Wavelets}) are proposed in the deep learning method, and relevant methods have been widely used in the fields of target detection, target segmentation, target recognition, etc, but no relevant research results have been found in the existing deep learning image fusion algorithm. The main reason is that in order to prevent the loss of image details in the process of image fusion, the pooling layer is not introduced. In this paper, on the one hand, we use the convolutional neural network model to extract the local and global features of the image, on the other hand, we use the visual attention model to introduce the global features to a certain extent.

To sum up, based on the previous study of human visual selection characteristics, the auxiliary learning mechanism and subjective attention characteristics of the human visual perception system, we proposed the image fusion theory of multi-task collaborative optimization guided by human visual subjective attention, and built an end-to-end unsupervised learning network framework based on our image fusion theory. Our proposed image fusion theory of human visual attention guidance can alleviate the problem of single loss of image evaluation and perception of image contextual awareness to a certain extent. It can effectively guide the network to carry out intentional learning, and has greater advantages in practical application compared with the guidance of single loss function. At the same time, aiming at the problem of the lack of global information of the existing image fusion features, we combine the advanced semantic information with the upsampling operation to achieve the effective fusion of global and local features. By introducing human subjective intention into the image fusion task and simulating the learning mechanism of human vision system, the image fusion effect can have dynamic context-awareness ability, and its fusion effect will be more robust.

The rest of this paper is laid out as follows. In section 2, we discuss the existing research work and problems related to image fusion theory. In section 3, we analyze our image fusion theory and build a top-down unsupervised learning image fusion framework guided by subjective visual attention. In section 4, experiments and results analysis of different algorithms on different public datasets. And the results are qualitatively analyzed and discussed in Section 5, and the experimental conclusions are summarized in Section 6.

\section{Proposed image fusion method}
\begin{figure}[ht]
	\setlength{\belowcaptionskip}{-0.5cm}
	\centering
	\includegraphics[width=1\textwidth]{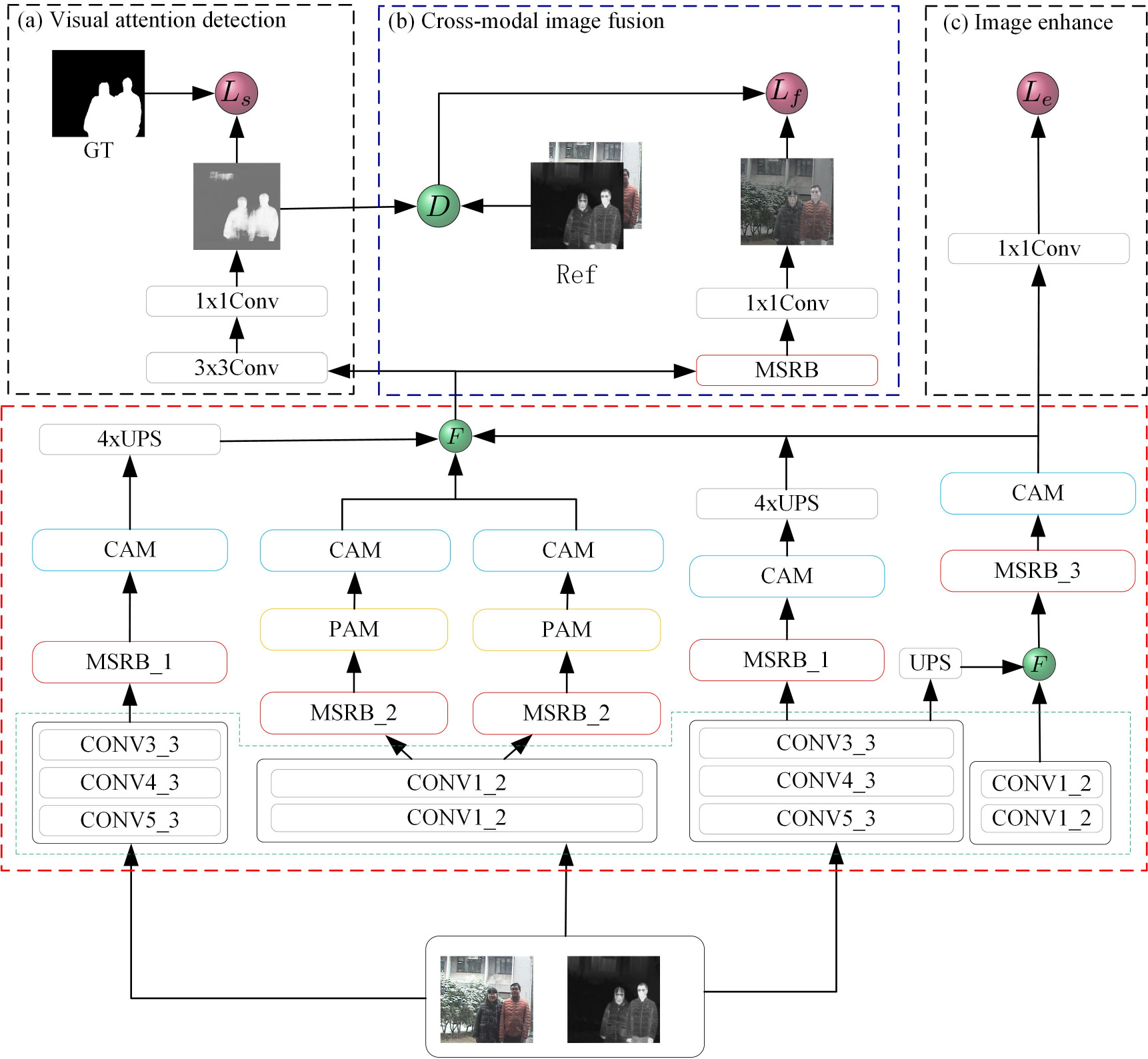}	
	\label{general}
	\caption{Multi-task image fusion framework guided by subjective visual attention. The visual attention detection sub-network detects the region of attention target or the region of human gaze through the attention fusion of multimodal images. The main network of cross-modal image fusion can guide the task of cross-modal image fusion through the output of subjective visual attention detection network, and make the main network of image fusion join the subjective intention of human beings. The image enhancement network can provide more hidden features which can not be captured by the main network for image fusion. MSRB \cite{LiJ.2018Mrnf}, CAM \cite{hu2017squeezeandexcitation}, PAM \cite{JaderbergM.2015Stn}, D and F indicate the multi-scale residual block, the channel attention module, the special attention module, the element-wise product, and the feature map concat operator, respectively. The module in the green frame corresponds to the weight of pre-trained model.}
	\label{general}
\end{figure}

As shown in \ref{general} , it is based on the image fusion theory proposed by us to build a top-down subjective attention guided multi-task assisted learning image fusion framework. The image fusion theory of multi-task auxiliary optimization of subjective visual attention guidance proposed by us needs to be completed in the following four steps: firstly, we proposed a detection and fusion method of cross-modal combined visual attention saliency map for cross-modal image fusion task; Then, we proposed the image fusion theory of visual attention guidance; secondly, we proposed a multi-task assisted learning optimization image fusion method. Finally, an end-to-end multi-task assisted learning image fusion network based on visual attention guidance is constructed.

\subsection{Local and global feature extration}
\begin{figure}[ht]
	\centering
	\includegraphics[width=1\textwidth]{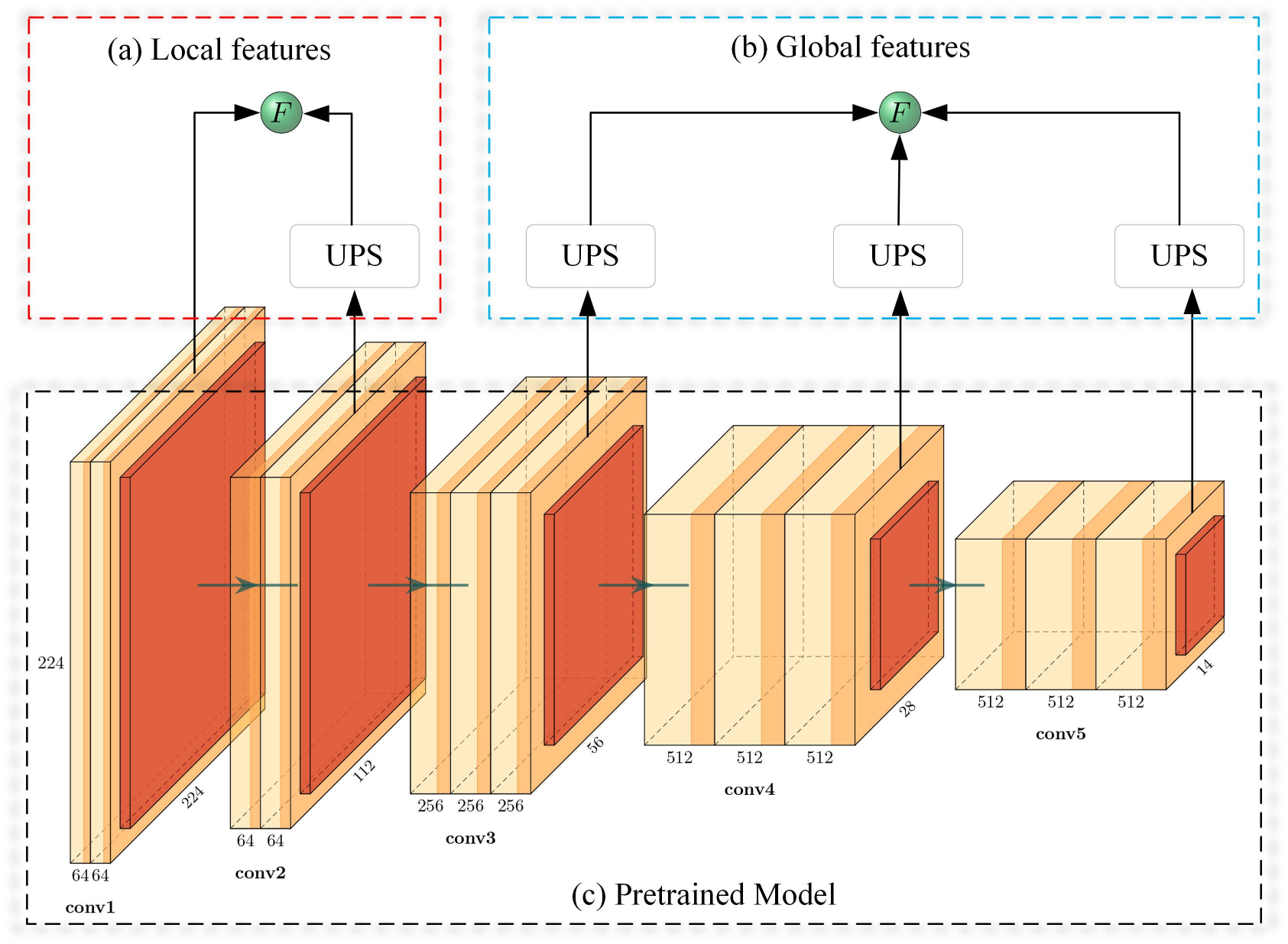}
	\caption{Local and global feature integration method. (a) indicates local features extraction process. (b) indicates global features extraction process.\label{light}}
	
\end{figure}

In the task of image fusion, the image fusion algorithm based on deep learning seldom considers the global features \cite{Lahoud2019FastZERO,Li_2018DL,Liu2017InfraredCNN,Yu2017medicalCNN,PrabhakarK.Ram2017DADU,MaBoyuan2019SAUD}, but the global features contain rich semantic information, which is as important as the local features. In order to solve this problem, we design a fusion module of global feature and local feature. As shown in Fig.\ref{light}, we first extract the low-level detail features and high-level semantic features from the pre-trained deep learning model, and then perform bilinear interpolation processing for the high-level semantic features, and then perform channel superposition and fusion with the low-level detail features.

As shown in Fig. \ref{general}, it is the global feature and local feature extraction module proposed by us. The mathematical model is shown in Eq. (\ref{gs1}):

\begin{equation}
\begin{array}{l}{F_{\text {low}}=P_{\text {local}}^{i}\left(I_{j}\right)+P_{\text {local}}^{i+1}\left(I_{j}\right) * U P S_1} \\ {F_{\text {high}}=\sum_{i=3}^{5} P_{\text {high}}^{i}\left(I_{j}\right) * U P S_{i}}\end{array}
\label{gs1}
\end{equation}

In the Eq. \eqref{gs1}, $F_{low}$ represents the bottom local feature of the fusion, which mainly includes the corner, edge, texture and other details of the image. $F_{high}$ represent the high-level semantic features of fusion. $P_{\text {local}}^{i}$ represents the local features of the i-th convolution layer of the pre-trained model.  represents the top adoption operation of high-level semantic feature map. In this paper, we mainly use the feature map of the first and second layers of the pretrained deep learning model (VGG16, VGG19 and Resnet101, etc.) as the bottom local feature, and use the feature map of the third, fourth and fifth layers as the global feature.

\subsection{Mul-task Auxiliary Learning mechanism}
In the task of cross-modal image fusion, there are two primary problems. Firstly, there are few ground truth labels in cross-modal data set, so it is impossible to carry out supervised learning training; secondly, due to the difference of imaging attributes, cross-modal data lacks perfect image quality evaluation loss function. To a great extent, this restricts the practical application of cross-modal image fusion technology. According to the research of neuroscience and cognitive psychology \cite{MillerCPortex,LiuDing2014Mpad}, the prefrontal cortex of human brain has the ability of working memory and dynamic situational learning, that is, when dealing with new tasks, it often guides the new tasks according to the experience model established by the existing tasks, establishes the cognition of the new tasks, so as to achieve rapid learning. At the same time \cite{GuangYang2009Smds} theoretical research shows that when facing new tasks, people are more likely to modify the existing deep neural network and add parameters, rather than establish a new network for learning every time they encounter new tasks. The two are not independent. It is the human multi-task auxiliary learning mechanism that ensures that human beings can learn quickly when facing new tasks. Therefore, in order to narrow the gap between the image fusion task and human, we introduce the human brain auxiliary learning mechanism in the image fusion task, and optimize the main task of image fusion through multi-task auxiliary learning. This method not only effectively avoids the problem of imperfect loss function of image quality evaluation, but also introduces humans subjective visual attention through multi-task auxiliary learning mechanism to guide the network learning \cite{fang2019crossmodal}.

In order to express the structure of the network more clearly, we build a mathematical model \cite{fang2019crossmodal}, which is shown in Eq. \eqref{gs7}.

\begin{equation}
\label{gs7}
Task_{i}^{(l)}=R\left(W_{i-1}^{(l)} x_{i-1}^{(l)}+\sum_{j<l} W_{i-1}^{(j)} x_{i-1}^{(j)}\right)
\end{equation}
where $Task_{i}^{(l)}$ represents task $0$-$l$ at layer i; R represents the nonlinear activation layer, $W_{i-1}^{(l)}$ represents the convolution weight of task $l$ in layer $i-1$ network; $x_{i-1}^{(l)}$ represents the input of task $l$ in the $i-1$ network; $\sum_{j<l} W_{i-1}^{(j)} x_{i-1}^{(j)})$ represents the sum of $l-1$ tasks in layer $i-1$ convolution neural network \cite{fang2019crossmodal}.

In the main task of image fusion, we use structure similarity index loss \cite{1284395,Li2018DenseFuse}, perception loss \cite{JohsonJustin2016PLfR} and edge loss \cite{JohsonJustin2016PLfR}.
\begin{equation}
L_{\text {f}}=L_{\text {SSIM }}+L_{Perceptual}+L_{\mathrm{Edge}}
\end{equation}
SSIM loss represents the similarity measurement of brightness, contrast and structure between the predicted image and the reference image. The higher the index, the higher the image similarity. The SSIM losses are shown in Eq. \eqref{gs6} .
\begin{equation}
\label{gs6}
{\text L_{SSIM(x,y)}}=1-\sum_{x, y} \frac{2 \mu_{x} \mu_{y}+C_{1}}{\mu_{x}^{2}+\mu_{y}^{2}+C_{1}} \cdot \frac{2 \sigma_{x} \sigma_{y}+C_{2}}{\sigma_{x}^{2}+\sigma_{y}^{2}+C_{2}} \cdot \frac{\sigma_{xy}+C_{3}}{\sigma_{x} \sigma_{y}+C_{3}}
\end{equation}
where $\mu_{x}$ and $\mu_{y}$ represent the mean of x and y. $\sigma_{x}$, $\sigma_{y}$ represent represent the standard deviation of x and y.

The perceptual loss function is mainly proposed to overcome the image smoothing and blur caused by MSE loss. The loss function uses advanced semantic information to solve MSE loss. The perceptual loss are shown in Eq. \eqref{gs10}.
\begin{equation}
\label{gs10}
L_{Perceptual}=\frac{1}{C*H*W}\left\|\phi({I_{predict}})-\phi(I_{reference})\right\|_{2}^{2}
\end{equation}

The edge loss function \cite{ZhaoTing2019PFAN} is mainly used to learn the edge information of the image to be fused. The edge loss can be obtained by convolution of the predicted image and the reference image by the second-order differential Laplacian operator. On the basis of the edge image, the binary cross entropy (BCE) loss or SSIM loss can be performed. The edge loss combined with SSIM loss are shown in Eq. \eqref{gs11}. The edge loss combined with BCE loss only needs to replace SSIM loss function.
\begin{equation}
\label{gs11}
L_{Edge}= L_{SSIM}(f_{laplace}(I_{predict},k),f_{laplace}(I_{reference},k))
\end{equation}
where $f_{laplace}$ represent Laplace function, k represents Laplace convolution kernel.

In our proposed image fusion theory, there are two subtasks, one is top-down visual attention target region detection task, the other is image enhancement task. The top-down visual attention detection task is mainly used to detect the significant areas related to the subjective task, which is the basis for the subsequent image fusion. The image enhancement task is mainly used to enhance the local features of the objects concerned by human subjective intention. The loss function of the two sub-tasks is the same as that of the main task.

\subsection{Image fusion}
In order to make the effect of image fusion more in line with the subjective attention of human beings, we proposed an image fusion theory guided by the subjective attention of human vision. Humans subjective attention is mainly used to guide the network model to extract the most relevant information from a large number of data to be fused and give different display weights to the data with different relevance of tasks. By giving a large display weight to the current region of concern, we can effectively avoid the distraction of human beings attention from the significant information that is not helpful to the current task.

\subsubsection{Cross-modal visual attention fusion detection method}
In our algorithm framework, we mainly use the top-down task-related attention model, mainly because the bottom-up significance model can distract human beings task-oriented subjective attention to a certain extent, and the top-down visual attention model is affected by human beings own emotions, wills and external stimuli. In order to maintain the high attention of the landing mission, we try to avoid the feature fusion operation from the bottom-up significant region. In the existing top-down visual attention detection methods, more attention saliency map detection of single modal data \cite{WangW.2018SODD}, and less cross-modal visual attention detection. Therefore, the existing attention detection model has a good performance in a specific data set, while when migrating to other modal data, the problem of false detection or missing detection often occurs in the region of attention. To solve this problem, a theory of attention detection based on cross modal image fusion is proposed As shown in Fig. \ref{general}, it is our attention detection network based on the attention detection theory of cross-modal image fusion.

In the top-down attention extraction network proposed by us, the cross-modal attention fusion module can switch according to different data tasks, including summation, weighted average, nonlinear to maximum and other fusion criteria. As shown in Eq.\ref{gs2}, it is the physical model of our attention fusion.

\begin{equation}
F_{S(x, y)}=\sum_{i=1}^{W * H}\left(\alpha_{i} I_{j}(x, y)+\beta_{i} I_{j+1}(x, y)\right)
\label{gs2}
\end{equation}

In the Eq.\ref{gs2}: $S_{i}\left(I_{j}, I_{j+1}\right)$ represents the visual significance map of the i-th input image and $\alpha_{i}$ and $\beta_{i}$  represents the fusion weight of the two images at the pixel $(x, y)$. This parameter determines different fusion criteria, which can be specified manually or learned by learning optimization method. In this paper, we use the nonlinear and sum fusion criterion, which is obtained by deep learning network training.

\subsubsection{Image fusion theory of visual attention guidance}
\begin{figure}[ht]
	\centering
	\includegraphics[width=0.6\textwidth]{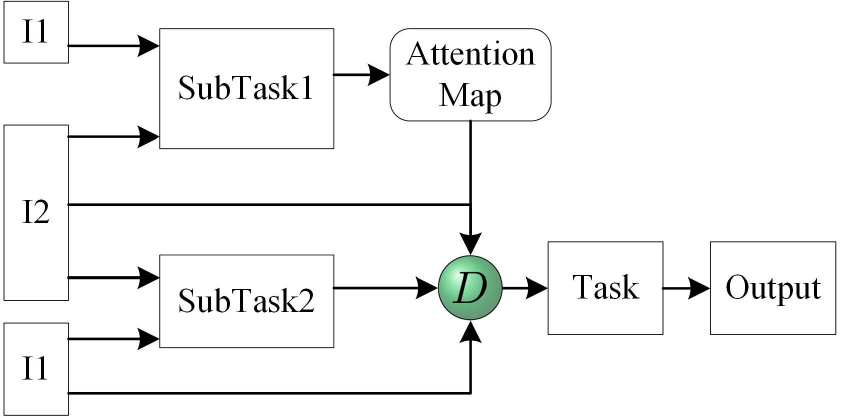}
	\caption{The multi-task auxiliary learning process of attention guidance.}
	\label{enf}
\end{figure}
In view of the problems existing in the existing image theory, a top-down subjective attention guided image fusion method is proposed. The method first performs subjective attention detection on images $I_j$ and $I_{j+1}$ of different modalities and gives different display weights to the attention region according to the relevance of the task. After the subjective attention image is fused, the subjective attention saliency map and the original image and the enhanced feature map are separately subjected to dot multiplication operations, and the results of the dot multiplication are used as the input of the cross-modal image fusion main task for convolutional neural network training to get the final fusion map. The image fusion process is shown in Fig.\ref{enf}.

The visual attention feature maps of different modalities are fused according to the characteristics of the task, and the result of the fusion is multiplied with the deep feature map to obtain a feature map with human subjective visual characteristics. As shown in Eq. \eqref{gs12}, it is a mathematical model of image fusion guided by subjective attention.

\begin{equation}
\begin{array}{l}{F_{f(x, y)}=N\left(F_{j}*\left(I_{j}\right), F_{j+1}*\left(I_{j+1}\right), F_{S(x,y)}*F_{E n h a m e}\left(F_{j}, F_{j+1}\right)\right)} \\ {F_{j}=F_{S(x, y)} I_{j}} \\{F_{j+1}=F_{S(x, y)} I_{j+1}}\end{array}
\label{gs12}
\end{equation}

In the Eq.\ref{gs12}, $F_{f(x, y)}$ represents the final result of image fusion; N indicates the deep learning weight model for image fusion. As shown in Fig.\ref{enf}, it is a schematic diagram of the principle of visual fusion guided image fusion proposed by us.

\subsection{Unsupervised Attention network}
As shown in Fig.\ref{general}, we build an unsupervised learning network framework based on the image fusion theory guided by human subjective attention. The network framework mainly includes one main task and two sub-tasks. The cross-modal image fusion task is the main task, which uses an end-to-end unsupervised learning network. In the main task network, our network is mainly composed of global feature and local feature extraction module, channel attention module, multi-scale feature extraction module and feature fusion module. In the main task training process, we used the original CVS image data set of 2,522, with a single resolution of 1280x1024, and expanded it to 7,566 through data enhancement. In order to increase the diversity of data, we added another 1800 pre-registered infrared and visible data set, with a single resolution of 640x480, and expanded it to 5,400 through data enhancement. In the main task training, all our images are input in the form of gray-scale image, and the image size is 256x256.

Sub-task 1 is a cross-modal subjective attention target detection task that uses an end-to-end supervised learning network. In this network, we introduce a channel attention module and a spatial attention module, which are used to select channel features and regional features. Our network is different from existing networks in that our attention detection network combines the advantages of different modal image data and effectively integrates features of different scales, global features, and local features. On the training data set, we used 1800 infrared and visible image data sets and 2522 CVS image data sets. These data sets include aligned original images and label images with subjective attention. But because it is cross-modal image data, neither of these two data sets has corresponding original ground truth labels. Through data augmentation, we obtained 12,966 training data sets and 5000 test data sets. Due to the limitation of computer memory, we adjusted the pre-processed image size to 256x256.

In order to enhance the visual attention region, we introduce sub-task 2 which is an end-to-end image enhancement network. In our image enhancement network, we use a stack self-encoding network to encode and decode images. The difference between this network framework and existing self-encoding networks is that we add densely connected modules and channels Attention module. In the sub-task 2 training phase, we used more than 70,000 training sets and 10,000 validation sets on the COCO2014 dataset. Due to the limitation of video memory, we adjusted the pre-processed image size to 256x256 \cite{fang2019crossmodal}.

In order to avoid the main task loss function affecting the convolution weights of the sub-tasks, we first train the sub-tasks separately and fix the sub-task convolution weights. The output of the subtask is taken as a part of the relevant node of the main task. 

\section{Experiments}
\subsection{Experiments Setup}
\label{setup}
In order to evaluate the robustness and generality of our algorithm, we performed relevant experimental evaluations on the CVS image data set, infrared and visible image data sets. First, the experimental comparison between our proposed image fusion algorithm framework and the existing algorithm framework in the CVS image data set, infrared and visible image data set. Then, we will evaluate the image quality of the two image data sets subjectively and objectively, and analyze the image data in detail.

In the first experiment, we first carried out experiments on the CVS image data set, which has 4000 pairs of original images. Secondly, we obtain infrared and visible images of natural scenes from RGBT-Saliency dataset \cite{WangG.2018Rsdb}, which includes 821 pairs of infrared image, visible images and ground truth labels. In all experiments, we transform all images into gray-scale images for subsequent image fusion training. We will compare experiments with 18 mainstream algorithms such as fast-zero-learning (FZL) \cite{Lahoud2019FastZERO}, deep learning (DL) \cite{Li_2018DL}, generative adversarial network for image fusion (FusionGAN) \cite{Ma2018Infrared}, laplacian pyramid (LP)
 \cite{Burt1987TheLP}, dual-tree complex wavelet transform (DTCWT) \cite{Liu2015MultiDSIFT}, multi-scale transform and sparse representation(LP-SR)
\cite{Liu2015ALPSR}, dense sift
(DSIFT) \cite{Liu2015MultiDSIFT}, convolutional neural network (CNN) \cite{Liu2017InfraredJSR-SD}, curvelet transformation (CVT)
\cite{Nencini2007RemoteCVT}, bilateral filter fusion method (CBF) \cite{Shreyamsha2015ImageCBF} , cross joint sparse representation
(JSR) \cite{Zhang2013Dictionary}, gradient transfer fusion (GTF) \cite{Ma2016InfraredGTF}, a ratio of low pass pyramid (RP) \cite{Toet1989ImageRP}, wavelet \cite{Chipman1995Wavelets}, IFCNN \cite{ZhangYu2020IAgi}, OURS, OURS+. In order to enhance the comparability of different image fusion algorithms, whether to add subjective attention is shown separately. The penultimate image fusion method is an image fusion effect that combines global features, local features, channel attention mechanism and image enhancement features. On this basis, the last image fusion method adds the fusion effect of human subjective attention mechanism. These algorithms have already published their code, and the relevant algorithm parameters are the same according to the settings in the public paper, and our paper-related procedures and data will then be published on github. For our proposed algorithm, we also conducted a comparative experiment on whether there is a channel attention module or not. Our experimental platform is desktop 3.0GHZ i5-8500, RTX2070, 32G memory \cite{fang2019crossmodal}.
 
 \subsection{Image fusion experiment of different data sets}
 
\subsubsection{CVS image fusion experiment}
\begin{figure}[ht]
	
	
	\centering
	\includegraphics[width=1\textwidth]{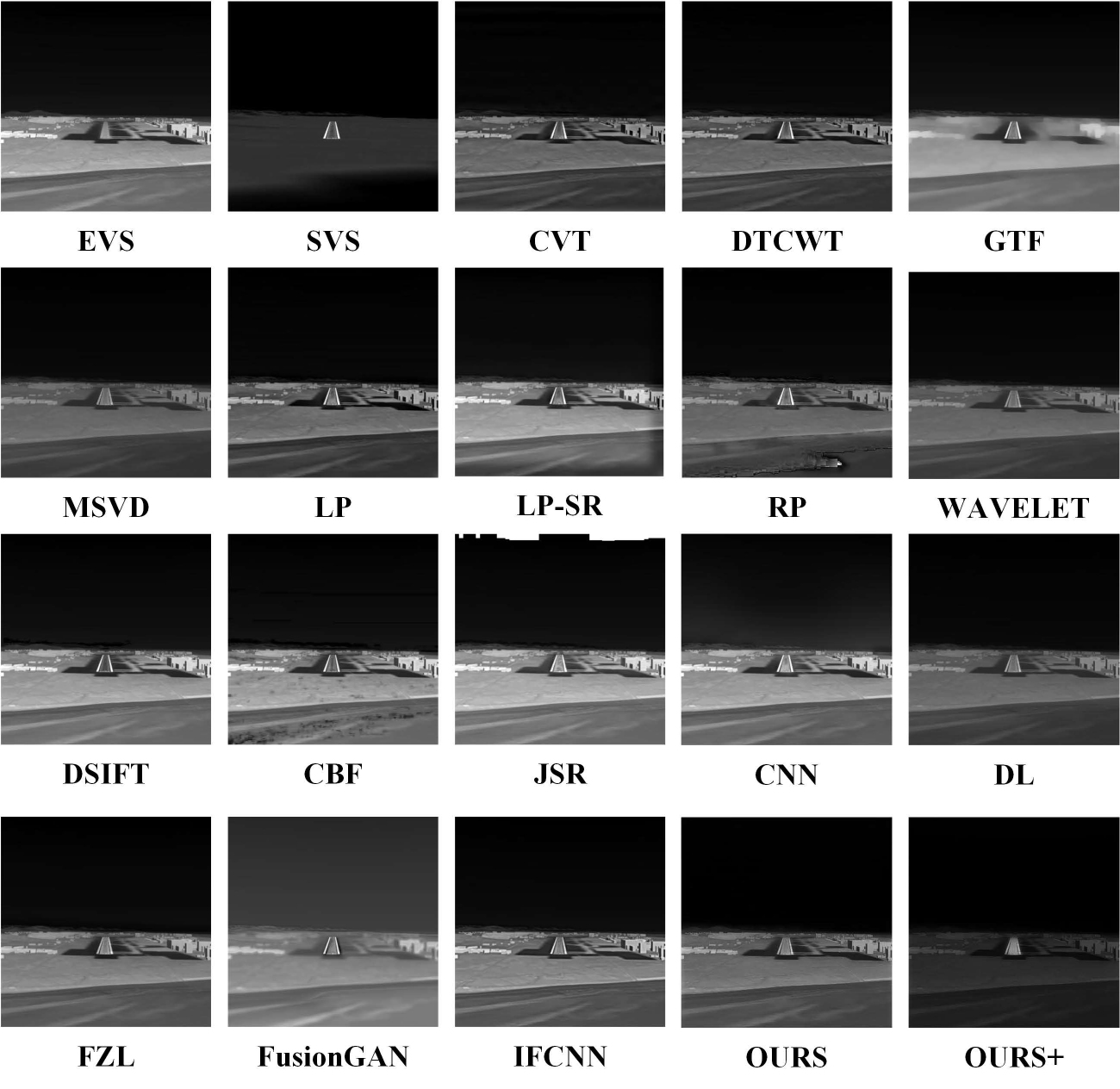}
	\caption{Qualitative fusion results on EVS and SVS images by different method. From left to right: CVT\cite{Nencini2007RemoteCVT},
		DTCWT\cite{Liu2015MultiDSIFT},
		GTF\cite{Ma2016InfraredGTF}, 
		MSVD\cite{Naidu2011Image}, 
		LP\cite{Burt1987TheLP}, 
		LPSR\cite{Liu2015ALPSR}, 
		RP\cite{Toet1989ImageRP}, 
		Wavelet\cite{Chipman1995Wavelets}, 
		DSIFT\cite{Liu2015MultiDSIFT},
		CBF\cite{Shreyamsha2015ImageCBF},
		JSR\cite{Zhang2013Dictionary},
		CNN\cite{Liu2017InfraredCNN},  
		DL\cite{Li_2018DL},  
		FZL\cite{Lahoud2019FastZERO}, 
		FusionGan\cite{MaFusionGAN}, 
		IFCNN\cite{ZhangYu2020IAgi}, 
		OURS,
		OURS+.}
	\label{f5}
	
\end{figure}

\begin{figure}[ht]
	
	
	\centering
	\includegraphics[width=1\textwidth]{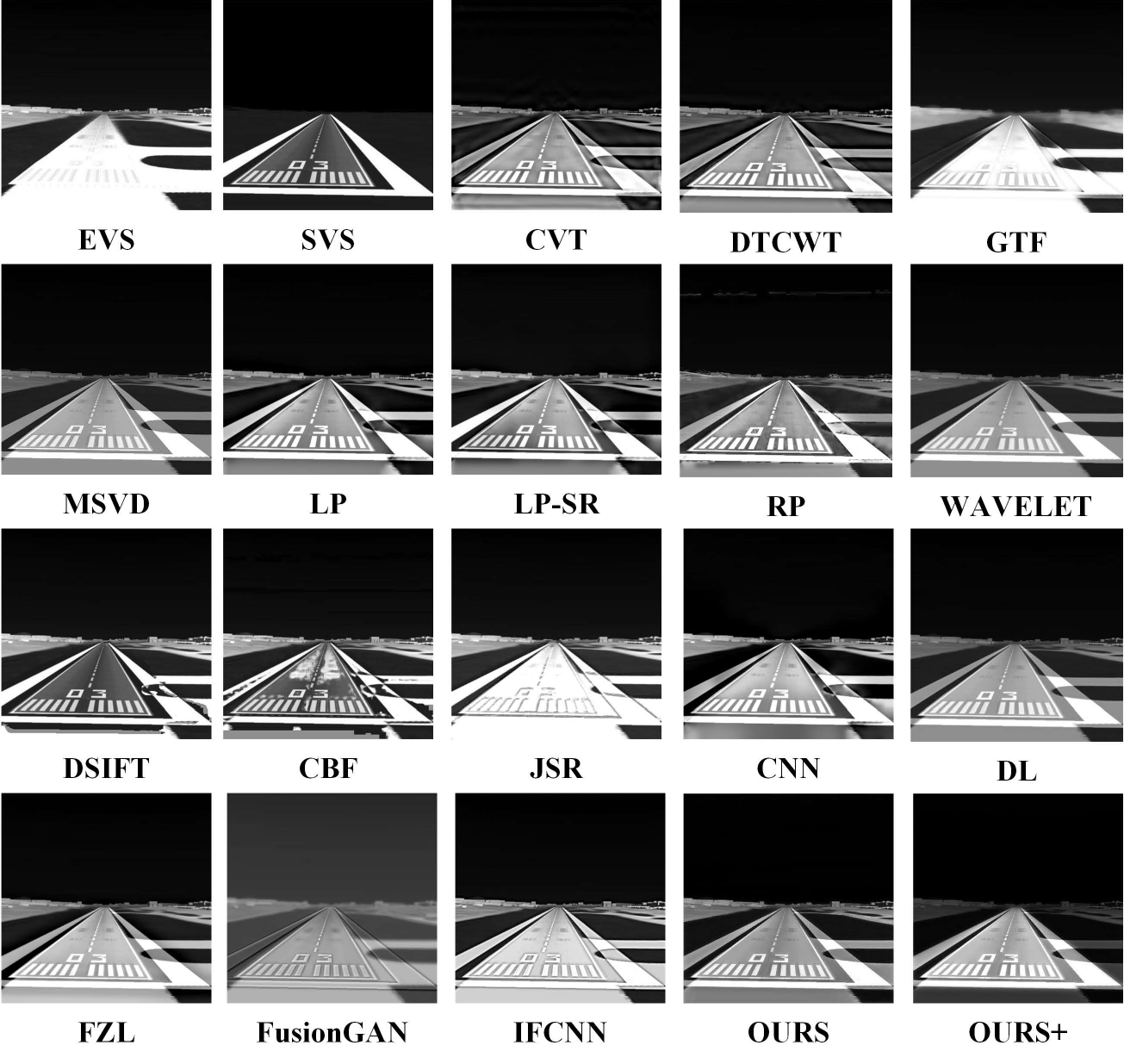}
	\caption{Qualitative fusion results on EVS and SVS images by different method. From left to right: CVT\cite{Nencini2007RemoteCVT},
		DTCWT\cite{Liu2015MultiDSIFT},
		GTF\cite{Ma2016InfraredGTF}, 
		MSVD\cite{Naidu2011Image}, 
		LP\cite{Burt1987TheLP}, 
		LPSR\cite{Liu2015ALPSR}, 
		RP\cite{Toet1989ImageRP}, 
		Wavelet\cite{Chipman1995Wavelets}, 
		DSIFT\cite{Liu2015MultiDSIFT},
		CBF\cite{Shreyamsha2015ImageCBF},
		JSR\cite{Zhang2013Dictionary},
		CNN\cite{Liu2017InfraredCNN},  
		DL\cite{Li_2018DL},  
		FZL\cite{Lahoud2019FastZERO}, 
		FusionGan\cite{MaFusionGAN}, 
		IFCNN\cite{ZhangYu2020IAgi}, 
		OURS,
		OURS+.}
	\label{f66}
	
\end{figure}

 As shown in Fig. \ref{f5},Fig. \ref{f66}, we can find our image fusion effect compared with the existing image fusion algorithm, it has a better subjective feeling, a clearer boundary, and no image fusion shake effect. In terms of image quality alone, the image fusion effect of the WAVELET, DL, FZL algorithms is better than the existing image fusion algorithms. Other image fusion algorithms, such as CNN, CVT, DTCWT, LP, etc., fuse well when the runway region is small, but when the runway region is large, an obvious image fusion shock effect will appear. Of course, image fusion operators such as CNN, DL, FZL, ifcnn have smaller image fusion noise than our image fusion effect, because on the one hand, the original image contains a lot of image noise, on the other hand, CNN, DL, FZL, ifcnn and other image fusion algorithms include image denoising algorithms, such as Guided Filtering (GF) \cite{Shutao2013ImageGF}, Contrast-Limited Adaptive Histogram Equalization (CLAHE) \cite{Zuiderveld:1994:CLA:180895.180940},etc. However, in the CVS image fusion task, it is not advisable to directly filter the EVS image, because the EVS image contains the real-time information of the airport runway. When it is far from the landing point, the size of the real-time dynamic information on the runway will be very small, sometimes it will be covered by noise data. The existing filtering algorithm will give the real-time dynamic information on the airport runway to Therefore, we can only filter SVS image data. Through the image fusion effect map of visual attention guidance proposed by us, our subjective attention can be focused on the runway region clearly, which effectively reduces the distraction of the non runway region to the pilot's attention.

 \subsubsection{Infrared and visible image fusion experiment}
 \begin{figure}[htp]
 	\centering
 	\includegraphics[width=1\textwidth]{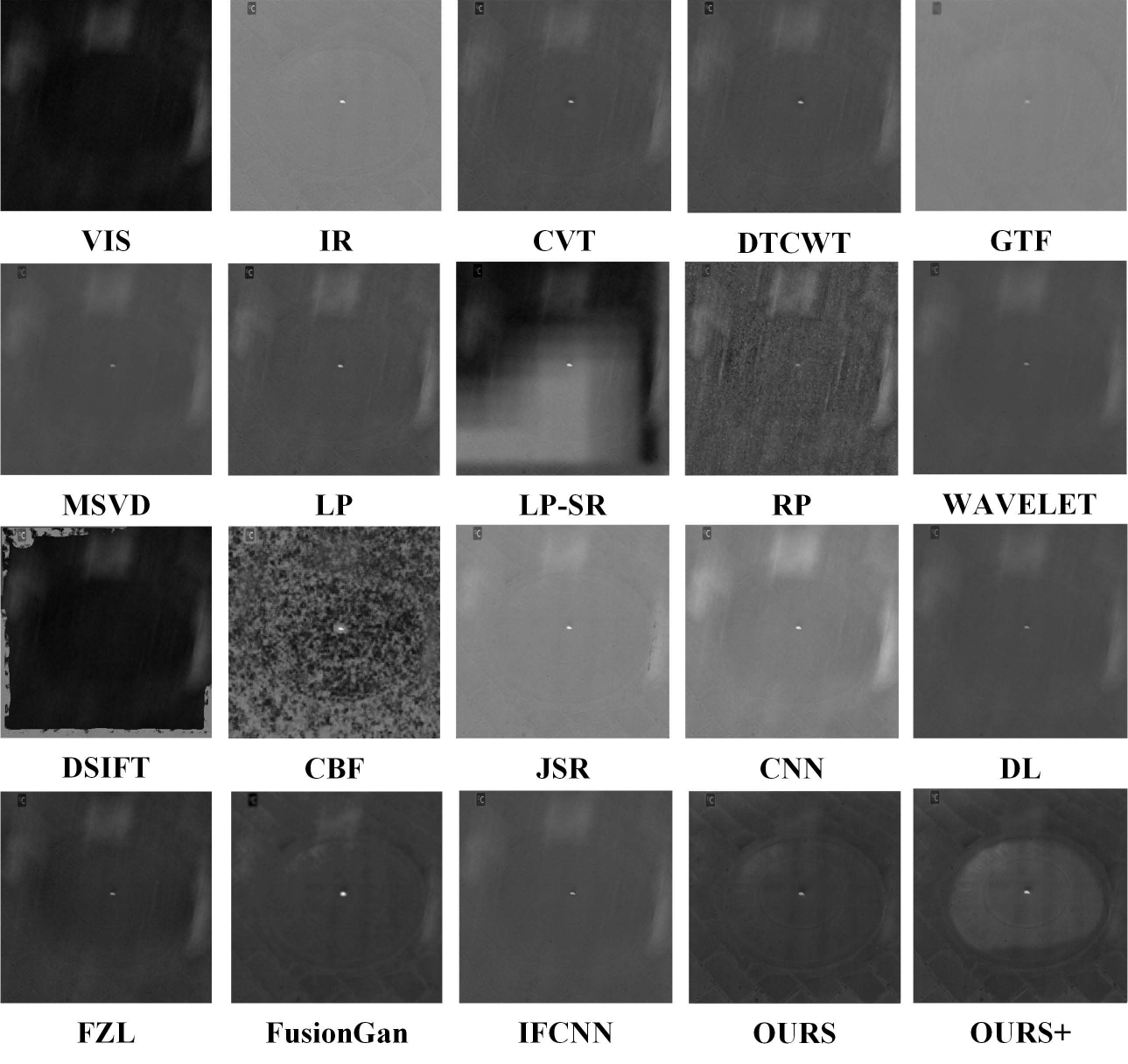}
 	\caption{Qualitative fusion results on visible and thermal infrared images by different method. From left to right : CVT\cite{Nencini2007RemoteCVT},
 		DTCWT\cite{Liu2015MultiDSIFT},
 		GTF\cite{Ma2016InfraredGTF}, 
 		MSVD\cite{Naidu2011Image}, 
 		LP\cite{Burt1987TheLP}, 
 		LPSR\cite{Liu2015ALPSR}, 
 		RP\cite{Toet1989ImageRP}, 
 		Wavelet\cite{Chipman1995Wavelets}, 
 		DSIFT\cite{Liu2015MultiDSIFT},
 		CBF\cite{Shreyamsha2015ImageCBF},
 		JSR\cite{Zhang2013Dictionary},
 		CNN\cite{Liu2017InfraredCNN},  
 		DL\cite{Li_2018DL},  
 		FZL\cite{Lahoud2019FastZERO}, 
 		FusionGan\cite{MaFusionGAN}, 
 		IFCNN\cite{ZhangYu2020IAgi}, 
 		OURS,
 		OURS+.}
 	\label{f6}
 	
 \end{figure}

\begin{figure}[ht]
	
	
	\centering
	\includegraphics[width=1\textwidth]{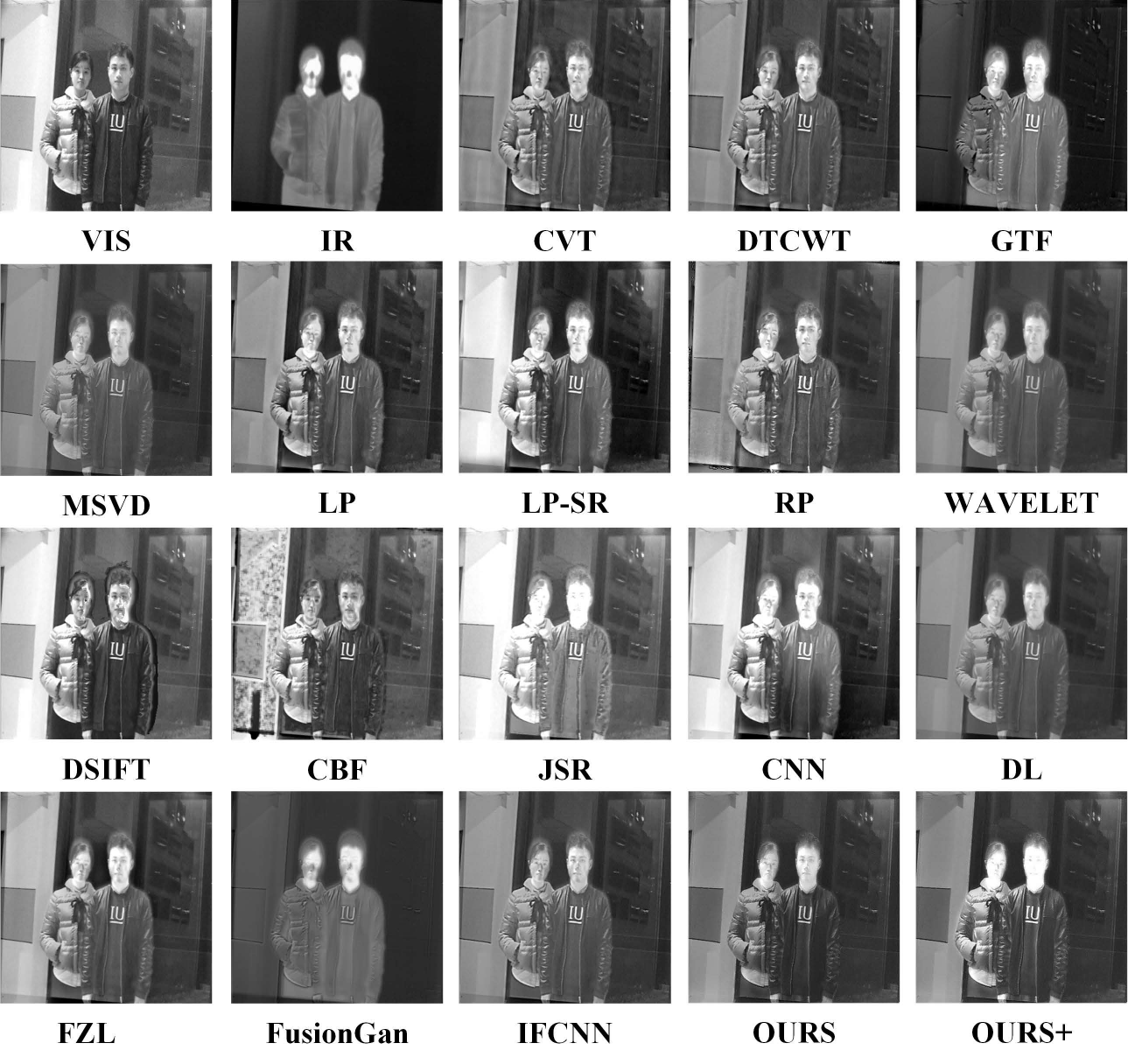}
	\caption{Qualitative fusion results on Visible and thermal infrared images by different method. From left to right: CVT\cite{Nencini2007RemoteCVT},
		DTCWT\cite{Liu2015MultiDSIFT},
		GTF\cite{Ma2016InfraredGTF}, 
		MSVD\cite{Naidu2011Image}, 
		LP\cite{Burt1987TheLP}, 
		LPSR\cite{Liu2015ALPSR}, 
		RP\cite{Toet1989ImageRP}, 
		Wavelet\cite{Chipman1995Wavelets}, 
		DSIFT\cite{Liu2015MultiDSIFT},
		CBF\cite{Shreyamsha2015ImageCBF},
		JSR\cite{Zhang2013Dictionary},
		CNN\cite{Liu2017InfraredCNN},  
		DL\cite{Li_2018DL},  
		FZL\cite{Lahoud2019FastZERO}, 
		FusionGan\cite{MaFusionGAN}, 
		IFCNN\cite{ZhangYu2020IAgi}, 
		OURS,
		OURS+.}
	\label{f7} 
\end{figure}

 From Fig.\ref{f6}, we can see that when the image quality is seriously degraded, the existing image fusion algorithms, whether the traditional image fusion algorithm, or the deep learning image fusion algorithm, or the image fusion method combining the traditional method and the deep learning method, the image fusion effect is very poor, but our image fusion theory even if it does not add human's subjective attention before that, the texture details of the image can still be recovered better. After adding human's subjective attention, we can find that our image fusion effect is more helpful for human's current search task. From Fig.\ref{f7}, we can see that when the image quality is great, the existing image fusion algorithms, whether the traditional image fusion algorithm, or the deep learning image fusion algorithm, or the image fusion method combining the traditional method and the deep learning method, the image fusion effect has been significantly improved, but there is a little difference compared with our image fusion effect. The existing image fusion algorithm has the fuzzy effect of image fusion boundary, but our image fusion boundary is very clear. After adding human's subjective attention mechanism, our image fusion effect fully retains the advantage information of both infrared and visible images. At the same time, there are several algorithms for multi-focus data sets in our comparison algorithm, such as DSIFT, IFCNN. Of course, there are some general image fusion algorithms in our comparison algorithms, such as IFCNN and FZL. In the case of poor original image quality, the performance of these algorithms is poor, but when the image quality is good, these algorithms will have better performance than other algorithms, but the fusion effect is not the best. For example, DSIFT, FZL, or IFCNN, these algorithms are all feature engineering artificial design image fusion processes, to a certain extent, they add human beings prior knowledge of the current task, so there will be some commonality in multiple data sets. However, different tasks of data sets have different data noise, which is not enough to make the best image fusion effect in the process of artificial design. Even if IFCNN algorithm, although the fusion criteria are specified manually, to a certain extent, human prior knowledge is added, but its fusion weight is fixed and does not have the learning ability of dynamic tasks, when the fusion data is complex, image fusion effect will also be poor. Through the combination of multi-task auxiliary learning mechanism and human subjective visual attention, we can effectively improve the dynamic contextual perception ability of image fusion.

\subsection{Fusion metrics}
\begin{figure}[ht]
	
	
	\centering
	\includegraphics[width=1\textwidth]{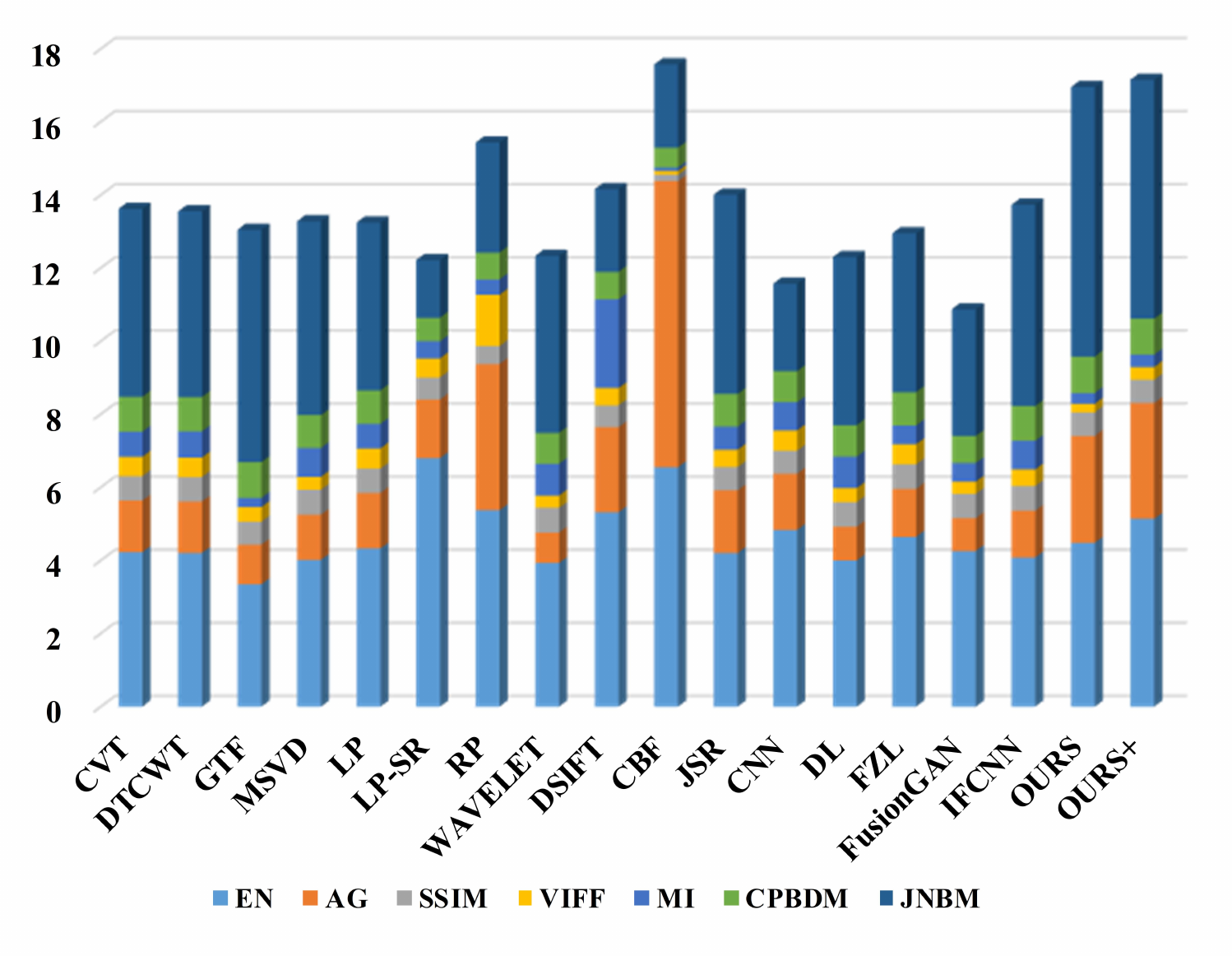}
	\caption{The first group of objective evaluation indexes of image fusion data.}
	\label{f66} 
\end{figure}
\begin{figure}[htp]
	
	
	\centering
	\includegraphics[width=1\textwidth]{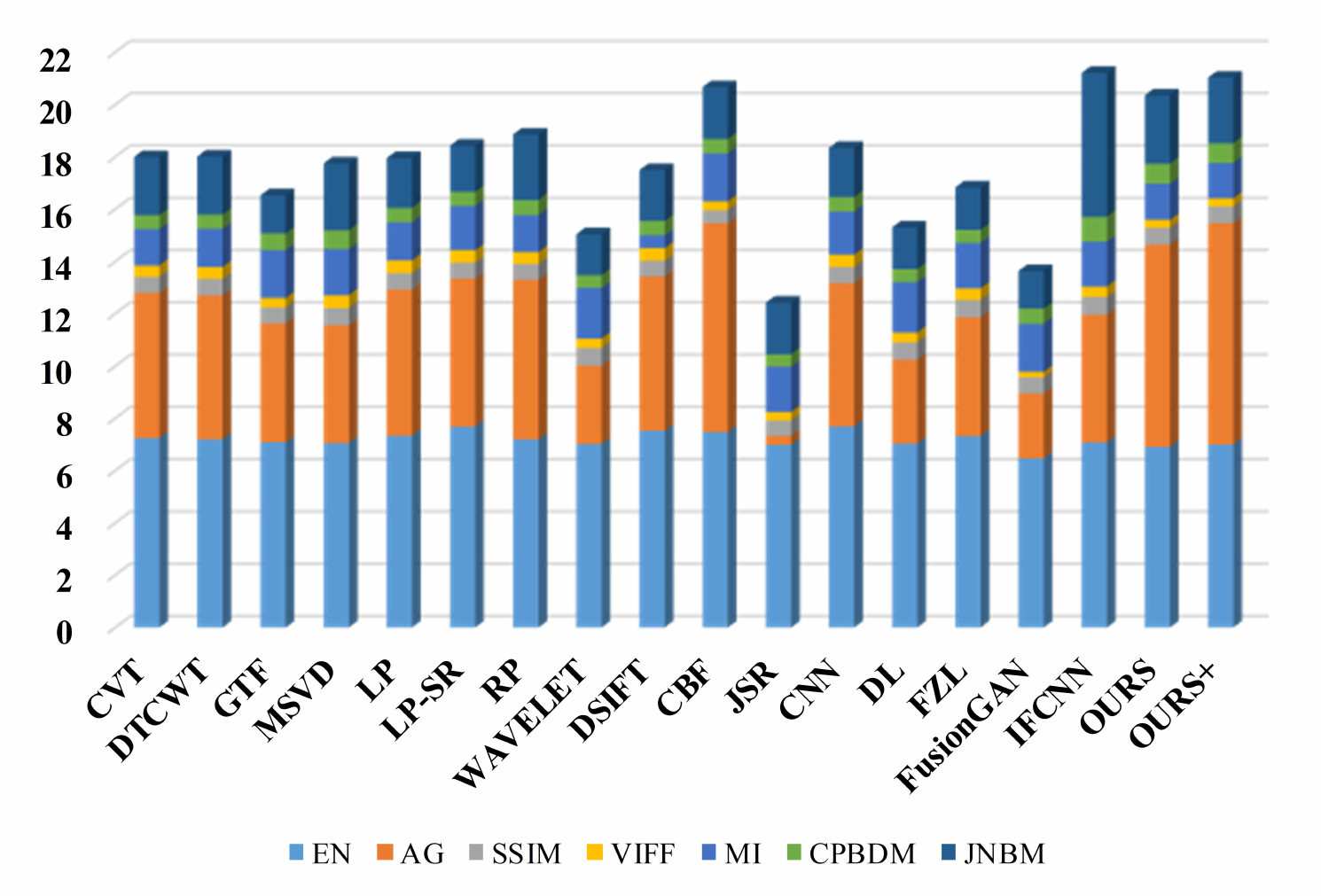}
	\caption{The second group of objective evaluation indexes of image fusion data.}
	\label{f77} 
\end{figure}

In order to qualitatively evaluate the performance of different algorithms, we mainly use six objective evaluation indexes of image: cumulative probability of blur detection (CPBD) \cite{NarvekarN.D2011ANIB}, just perceptible blur based on human vision (JNB) \cite{FerzliR2009ANOI}, visual information fidelity (VIF) \cite{Han2013A} , average gradient (AG) \cite{Cui2015Detail}, SSIM \cite{1284395}, mutural information (MI)
\cite{Qu2002Information}. We have carried out quantitative experiments on the infrared and visible image data set and CVS image data set.

From Fig.\ref{f66}, we can clearly find that almost all the indexes of our image fusion algorithm are not the best, which does not mean that our image fusion algorithm is not good. From Fig.\ref{f77}, we can also find that there are similar problems. In the second group of image fusion data, the effect of dsift and CBF image fusion algorithm is the worst, but their EN, AG, SSIM and MI are all very high. First of all, we analyze the image fusion effect corresponding to each image evaluation index. It is not difficult to find that although the relevant image quality evaluation index is very high, but their corresponding image subjective quality is very poor, such as LP-SR, CBF, MSVD, RP, GTF, etc. This is because there is still a big gap between the existing image quality evaluation indexes and the subjective visual evaluation methods of human beings. Secondly, SSIM, VIFF and MI indexes are based on the ground truth labels for quality evaluation. In this case, the closer the image fusion data is to the real reference image data, the greater the corresponding values of these three indexes will be. But in the task of cross-modal image fusion, due to the lack of real reference image, these three kinds of image quality evaluation indexes can not be used as the standard of image quality evaluation to some extent, and can not simply evaluate the image quality by the size of the three evaluation indexes. They are just a measure of the similarity between the fused image and cross-modal data. Although IFCNN algorithm has some advantages in the second group of image objective evaluation data, which is mainly caused by an abnormal value of its JNBM evaluation index, the overall image quality is still far from our fusion effect in terms of clarity.
\begin{figure}[ht]
	
	
	\centering
	\includegraphics[width=1\textwidth]{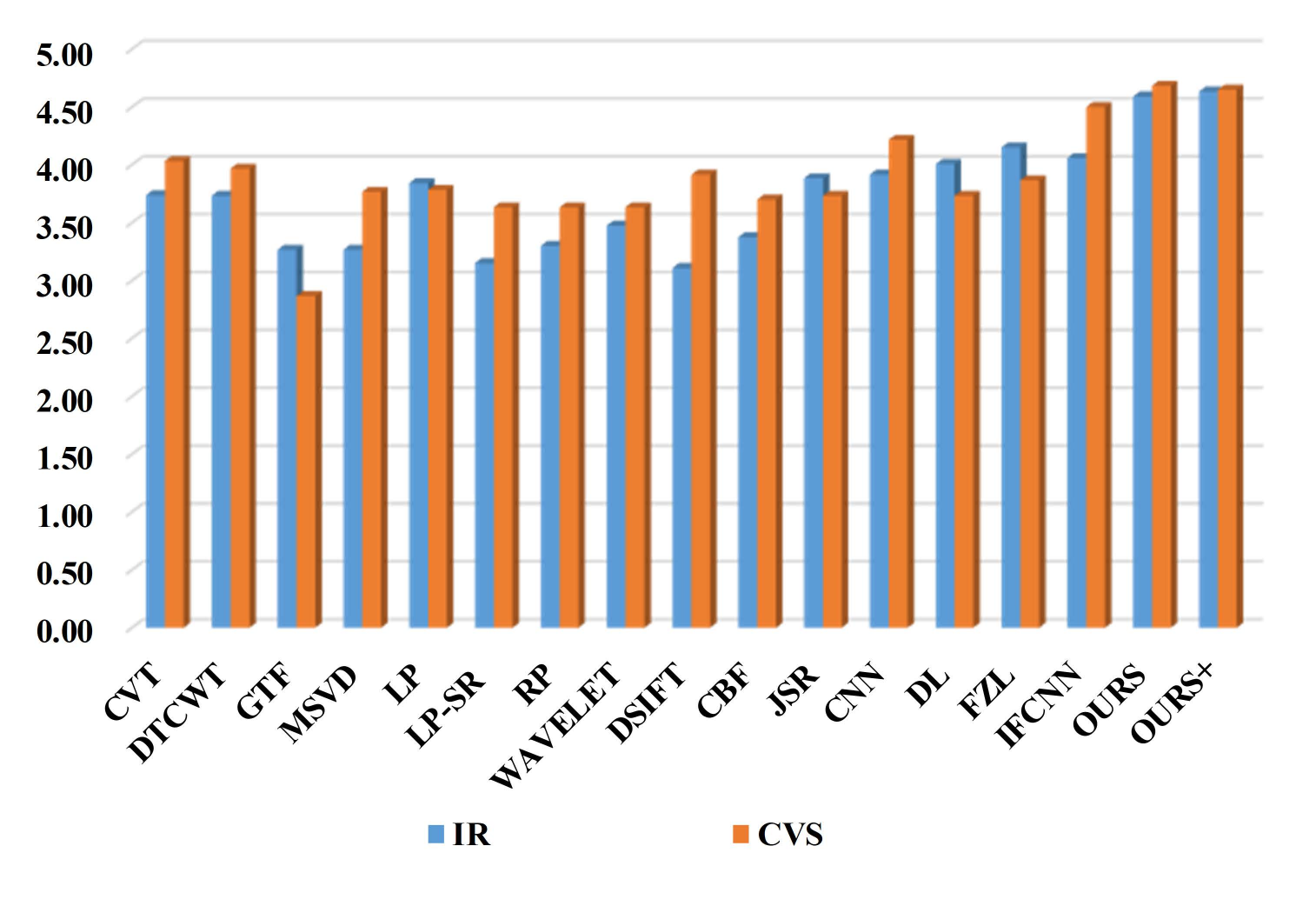}
	\caption{Comparison of MOS evaluation on different data set.}
	\label{Fig11} 
\end{figure}
On the one hand, in order to prove the superiority of our image fusion theory, we subjectively evaluate the fusion effect of 18 algorithms, and use mean opinion score (MOS) as the evaluation index of image quality. In this experiment, we invited 10 professors, doctoral researchers and other professional researchers of computer image processing to participate in the subjective quality evaluation of the fusion image of different algorithms. In the experimental data processing, we remove the highest and lowest scores of MOS, and take the average score of each data set test data as the final evaluation score. The experimental results are shown in Fig.\ref{Fig11}.

From Fig.\ref{Fig11}, we can find that our image fusion quality score is the highest in CVS image data set or infrared and visible image data set. This further proves the correctness of our proposed image fusion theory in the scheme and the accuracy of the above analysis.

\section{Discussion}
From a large number of experiments in the fourth chapter, it is verified that the image fusion theory guided by subjective visual attention proposed by us has stronger robustness and contextual awareness perception ability than the existing image fusion algorithms. We think there are several main reasons: first, the introduction of human visual subjective attention mechanism. Many existing image fusion theories take improving image quality as their ultimate goal. Due to the lack of subjective attention of human vision as guidance, and the imperfection of existing image quality evaluation indicators and the lack of real tags, this determines that the image fusion process is seriously lacking in contextual awareness perception, and the algorithm does not know the number of different It is more helpful for the current task to extract and fuse what kind of features. To solve this problem, compared with the traditional image fusion algorithm, deep learning method is particularly prominent. This is because the traditional algorithm is designed for a specific fusion task, so to a certain extent, the subjective intention of people is added. Although the deep learning algorithm has a strong advantage over the traditional algorithm in feature representation and relationship fitting, whether the deep network model can learn the features consistent with human subjective visual attention and whether the loss function is constructed reasonably is closely related. Therefore, human beings subjective visual attention can effectively guide the network to understand human intention and learn to learn for different tasks by combining the strong feature representation ability and nonlinear fitting ability of deep learning. Secondly, the auxiliary learning characteristics of human visual perception system. In the image fusion task, the existing image fusion algorithm based on deep learning has a serious dependence on the loss function on the one hand, and on the other hand, it is more of a single image fusion task deep learning method. In the task of cross-modal image fusion, the existing loss function of image fusion can not effectively guide the network to extract features and fit relationships, so we introduce the method of multi-task assisted learning. When single task network training and learning, it is often affected by data noise, insufficient training data, cross-modal and improper loss function, which leads to some hidden features of data can not be learned. Through auxiliary task learning, the learning ability of main task can be effectively optimized. In our network framework, both the reconstruction task and the visual attention detection task can be regarded as the sub-tasks of the main task of image fusion, and the experimental results also prove the effectiveness of the method. Finally, the combination of global features and local features. Global features often contain advanced semantic features of image, while local features contain more detailed texture information of image. The effective combination of the two can improve the representation ability of image.

\section{Conclusion}
Based on the robustness and contextual awareness of the human visual perception system, we proposed an cross-modal image fusion theory guided by human subjective visual attention. The biggest difference between our image fusion theory and current mainstream algorithms is: first of all, our image fusion theory is based on human subjective visual attention guidance, rather than the traditional image fusion theory. Our image fusion effect is more conducive to assist human decision-making in practical tasks. Secondly, the auxiliary learning mechanism is introduced into the image fusion task, which effectively optimizes the image fusion task and solves the image fusion problem caused by the loss function of image quality evaluation to a certain extent. Finally, the image fusion theory proposed in this paper is based on unsupervised learning method, does not need ground truth labels, and improves the universality of the algorithm to a certain extent. A large number of experiments show that our image fusion theory has stronger robustness and contextual awareness than the existing mainstream image fusion theory. Although our algorithm framework does not fully simulate human visual perception characteristics, our simulation of human visual perception characteristics is in line with the mechanism of the human visual system. In the task of cross-modal image fusion, although our image fusion theory has achieved relatively good results compared with the existing algorithms, there are still the following problems. First of all, we have only carried out experiments in the combination of CVS image data sets, infrared and visible image data sets, which need to be further extended to more image fusion tasks in the later stage. Secondly, in the image fusion task, we need to further deepen the research of working memory and contextual dynamic perception module, which is necessary for the future intelligent image fusion theory.

\begin{flushleft}	
\textbf{Acknowledgment}\\
\end{flushleft}
This work was supported by the National Natural Science Foundation of China under Grants nos. 61871326, and the Shanxi Natural Science Basic Research Program under Grant no. 2018JM6116, and National Natural Science Foundation of China under Grants nos. 61231016.

\section*{References}

\bibliography{mybibfile}

\end{document}